\documentclass{article}

\usepackage[preprint]{neurips_2024}

\usepackage[utf8]{inputenc} 
\usepackage[T1]{fontenc}    
\usepackage[hidelinks]{hyperref}       
\usepackage{url}            
\usepackage{booktabs}       
\usepackage{amsfonts}       
\usepackage{nicefrac}       
\usepackage{microtype}      
\usepackage[table,x11names, dvipsnames]{xcolor}         
\definecolor{myGreen}{RGB}{174, 225, 70}
\definecolor{myGreen}{RGB}{160, 225, 95}
\definecolor{myGreen}{RGB}{200, 238, 180}

\usepackage{microtype}
\usepackage{graphicx}
\usepackage{multirow}
\usepackage{array}

\usepackage{amsmath}
\usepackage{amssymb}
\usepackage{mathtools}
\usepackage{amsthm}
\usepackage{bm}
\usepackage{subcaption}

\usepackage{makecell}
\usepackage{multicol}
\title{Deep Vision-Based Framework for Coastal Flood Prediction Under Climate Change Impacts and Shoreline Adaptations}

\author{%
  Areg Karapetyan \quad Aaron C.H. Chow \quad Samer Madanat \\
  Division of Engineering\\
    New York University Abu Dhabi \\
    Abu Dhabi, United Arab Emirates\\
  \texttt{\{areg.karapetyan,cc6307,samer.madanat\}@nyu.edu} \\
}

\begin{document}

\maketitle

\begin{abstract}
In light of growing threats posed by climate change in general and sea level rise (SLR) in particular, the necessity for computationally efficient means to estimate and analyze potential coastal flood hazards has become increasingly pressing. Data-driven supervised learning methods serve as promising candidates that can dramatically expedite the process, thereby eliminating the computational bottleneck associated with traditional physics-based hydrodynamic simulators. Yet, the development of accurate and reliable coastal flood prediction models, especially those based on Deep Learning (DL) techniques, has been plagued with two major issues: (1) the scarcity of training data and (2) the high-dimensional output required for detailed inundation mapping. To remove this barrier, we present a systematic framework for training high-fidelity Deep Vision-based coastal flood prediction models in low-data settings. We test the proposed workflow on different existing vision models, including a fully transformer-based architecture and a Convolutional Neural Network (CNN) with additive attention gates. Additionally, we introduce a deep CNN architecture tailored specifically to the coastal flood prediction problem at hand. The model was designed with a particular focus on its compactness (only about 0.36 Mil. trainable parameters) so as to cater to resource-constrained scenarios and accessibility aspects. The performance of the developed DL models is validated against commonly adopted geostatistical regression methods and traditional Machine Learning (ML) approaches, demonstrating substantial improvement in prediction quality (ranging from 100\% to 400\% across most of the metrics). Lastly, we round up the contributions by providing a meticulously curated dataset of synthetic flood inundation maps of Abu Dhabi's coast produced with a physics-based hydrodynamic simulator, which can serve as a benchmark for evaluating future coastal flood prediction models.

\end{abstract}

\section{Introduction}
\label{sec:intro}

More than $60\%$ of the world's population resides in coastal areas within $60$ km of the shore\footnote{\href{http://www.oceansatlas.org/facts/en/}{United Nations Atlas of the Oceans}}. The looming global warming and its byproducts, such as rising sea levels\footnote{Approximately by $1$ m. in the worldwide mean by the end of the century \citep{IPCC}.} and more frequent and severe storm surges \citep{pnas.1703568114,sweet2022global}, render these regions increasingly susceptible to flooding incidents, which can inflict massive humanitarian, ecological, and economic devastation. To put this into perspective, in $2021$ alone, hydro-related disasters, including tidal and pluvial flooding and storm surges, caused worldwide more than \$ $224$ billion in loss and are projected to cost the global economy \$ $5.6$ trillion by 2050 \citep{reuters}. Ensuring the safety and longevity of coastal zones against rising floodwaters will require various interventions~\citep{Jongman2018}, among which construction of protective structures like seawalls, levees, and storm barriers is set to play a major role~\citep{Hummel2021EconomicFeedbacks}. Although beneficial for \textit{local} flood protection, these engineered fortifications may alter the hydrodynamics along the coast, amplifying water levels outside of their perimeter and spreading inundations to other (otherwise unaffected) regions~\citep{Wang2018TheInfrastructure,Haigh2020TheImplications, Hummel2021EconomicFeedbacks}. For instance, as demonstrated in~\citep{Hummel2021EconomicFeedbacks}, protection of certain individual stretches of the coast in San Francisco Bay can exacerbate flooding in other zones by up to $36$ million cubic meters owing to the effects of the modified shoreline geometry.

\begin{figure}[!b]
    \centering
    \begin{subfigure}{.33\textwidth}
    \includegraphics[width=\textwidth]{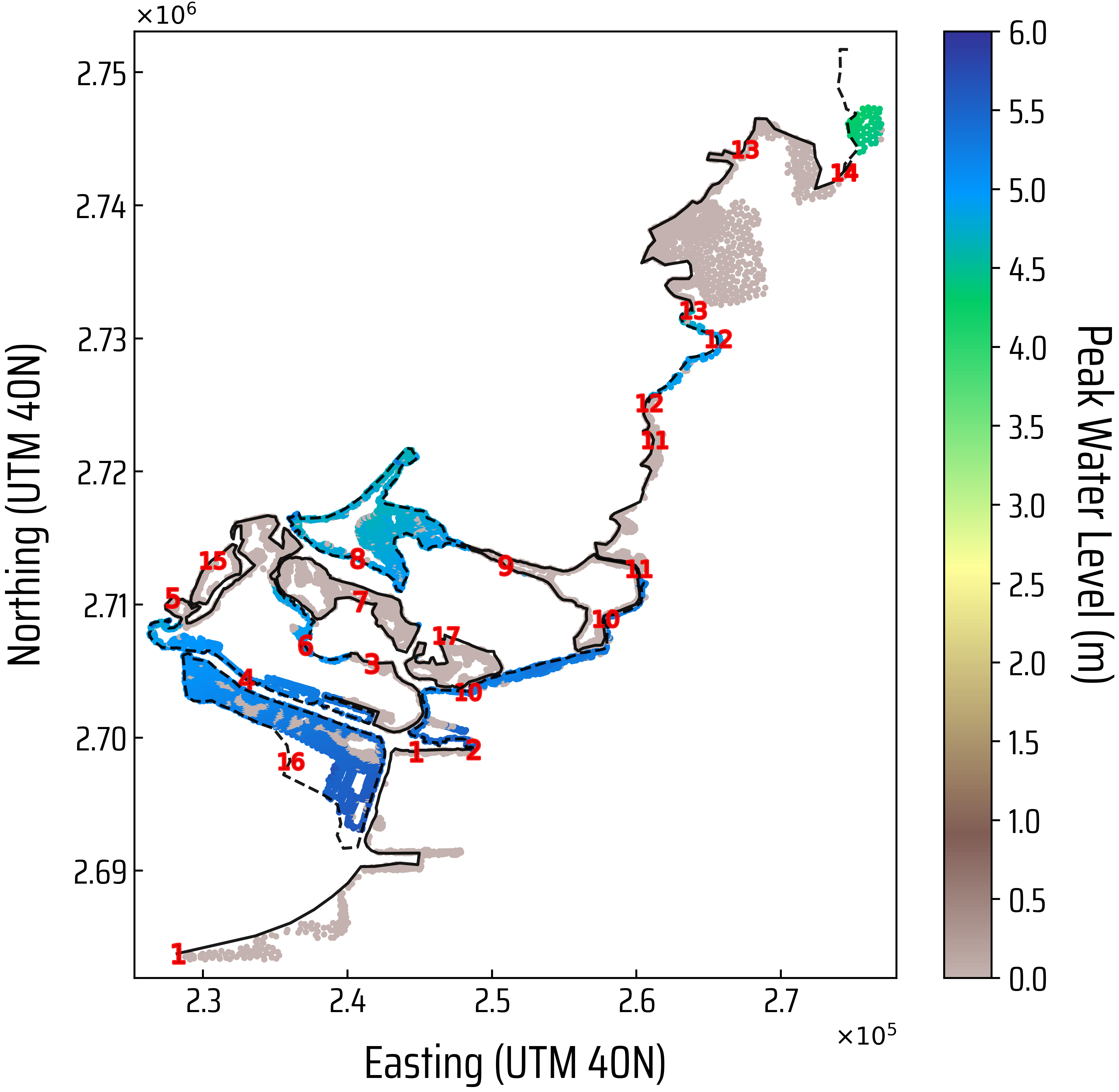}
    \caption{}
    \label{fig:fig1}
    \end{subfigure}
    \begin{subfigure}{.66\textwidth}
    \includegraphics[width=\linewidth]{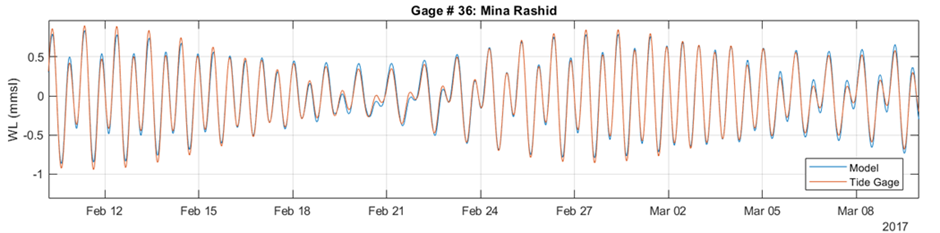}
    \includegraphics[width=\linewidth]{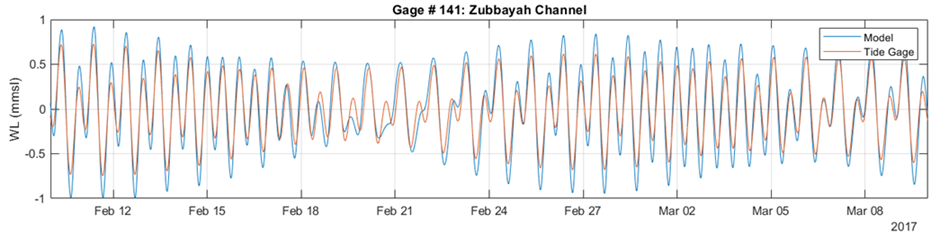}
    \caption{}
    \label{fig:fig2}
    \end{subfigure}
    \caption{ (A): Flood inundation map of Abu Dhabi's coast produced with a physics-based high-fidelity hydrodynamic simulator (presented in Sec.~\ref{sec:studyarea}) under a sample shoreline fortification scenario and SLR projection of $0.5$ meters. The coastline was partitioned into $17$ segments (precincts) of which those protected by seawalls are delineated by black solid lines whereas the unprotected are marked with dashed lines. The colorbar measures the magnitude of forecasted inundations at the nearshore locations of interest, which are plotted as scattered points. (B): Comparison of the simulated tidal model's output and TPXO8 Atlas' time-series data on water levels from 10 Feb to 10 March 2017 at Mina Rashid (on top) and Zubbayah Channel (on bottom). The units of vertical axes are in meters relative to the mean sea level.}
\end{figure}

Besides shoreline geometry, the factors governing coastal water levels include various natural and physical phenomena, such as tidal variations and forcing, seabed friction, winds, waves, and storms, which evolve and interact on different spatio-temporal scales. Progress in computing capabilities and numerical modeling techniques promoted the development of hydrodynamic simulators that can capture the aforementioned complex processes in a detailed and comprehensive manner. While these tools can accurately simulate nearshore hydrodynamics, providing fine-grade time-series data on estimated depth, duration, and velocity of floods, they are notoriously expensive in terms of computing time and resources. As a result, direct adoption of these high-fidelity simulators in studies requiring an extensive number of model realizations (e.g., optimal shoreline protection planning, sensitivity analysis) remains impractical. To exemplify, with such a hydrodynamic model in place (described in Sec.~\ref{sec:studyarea}), simulation and post-processing of \textit{a single} protection scenario for the coast of Abu Dhabi, illustrated in Fig.~\ref{fig:fig1}, demands approximately 72 hours on 128 HPC cores.

Data-driven supervised learning methods have emerged as promising alternatives that can unlock substantial speed-up gains. The concept behind these methods, which are also known as \textit{surrogate models} or \textit{metamodels} in the coastal engineering literature, is to emulate the high-fidelity numerical simulators by inferring the relationships between their inputs and outputs without explicit representation of the underlying sophisticated physical processes and the differential equations that govern their behavior. A prominent issue in the design of coastal flooding metamodels, as also highlighted in \cite{Jia2019InvestigationAnalysis, Kyprioti2021StormModels, rohmer2023improved}, concerns the dimensionality of the output predictions. The geographical domains considered in the analyses of nearshore areas are typically vast, leading to dense prediction tasks with tens of thousands of points. Another major obstacle for surrogate hydrodynamic models, especially those employing Deep Learning (DL) methods~\citep{stanton2023overcoming}, lies in the scarcity of training data since, as noted above, generating annotated samples via high-fidelity simulators consumes significant time and resources. These two factors combined have posed a solid challenge, leaving the development of performant DL-based surrogate models for high-resolution coastal flood prediction problems elusive.  

To reinforce the arsenal of coastal inundation metamodeling techniques, this paper presents a data-driven framework for synthesizing accurate and reliable DL-based coastal flood prediction models in low-resource learning settings. For comprehensiveness, the flood prediction problem is studied considering both anticipated climate change-induced effects and shoreline adaptations. The core idea behind the framework is to recast the underlying multi-output regression problem as a computer vision task of translating a two-dimensional segmented grid into a matching grid with real-valued entries corresponding to water depths. This, in turn, enables the incorporation of effective data augmentation techniques, thereby facilitating the training of high-performance DL-based surrogate models. In summary, the key contributions of the present work are three-fold:


\begin{itemize}
    \item First, we provide a systematic approach for training efficient Deep Vision-based end-to-end coastal flood prediction models in low-data settings. We test the proposed methodology on different neural networks, including two existing vision models: a fully transformer-based architecture (SWIN-Unet~\citep{cao2022swin}) and a Convolutional Neural Network (CNN) with additive attention gates (Attention U-net~\citep{oktay2022attention}). The developed models are benchmarked against existing geostatistical regression methods and traditional ML approaches commonly employed in the field. The comparison results reveal significant gains in predictive performance, with improvements from the developed DL-based surrogate models ranging from 100\% to 400\% across almost all metrics. The complete source code for reproducing the conducted analysis, including the trained models, can be accessed at \href{https://github.com/Arnukk/CASPIAN}{\color{blue}https://github.com/Arnukk/CASPIAN}.
    
    \item Next, we introduce a deep CNN architecture, dubbed Cascaded Pooling and Aggregation Network (CASPIAN), stylized explicitly for the coastal flood prediction problem at hand. The introduced model was designed with a particular focus on its compactness and practicality to cater to resource-constrained scenarios and accessibility aspects. Specifically, featuring as little as $0.36$ Mil. parameters and only a few main hyperparameters, CASPIAN can be easily trained and fine-tuned on a single GPU. On the current dataset, the performance of CASPIAN approached remarkably close to the results produced by the physics-based hydrodynamic simulator (on average, with 97\% of predicted floodwater levels having less than 10 cm. error), effectively reducing the computational cost of producing a flood inundation map from \textit{days to milliseconds}.

    \item Lastly, we round up the contributions by providing a carefully curated database of synthetic flood inundation maps of Abu Dhabi's coast under $174$ different shoreline protection scenarios. The maps were generated via a high-fidelity physics-based hydrodynamic simulator (see Sec.~\ref{sec:studyarea}) under a 0.5-meter SLR projection. The provided dataset, which is available at \href{https://doi.org/10.7910/DVN/M9625R}{\color{blue}https://doi.org/10.7910/DVN/M9625R}, to the best of our knowledge, is the first of its kind, and thus can serve as a benchmark for evaluating future coastal flooding metamodels.
\end{itemize}

\if
Our
models are flexible in terms of model size, and can have as
little as 0.28M parameters while achieving competitive results. Our best model can reach 98\% accuracy when training from scratch on CIFAR-10 with only 3.7M parameters,
which is a significant improvement in data-efficiency over
previous Transformer based models being over 10x smaller
than other transformers and is 15\% the size of ResNet50
while achieving similar performance. CCT also outperforms many modern CNN based approaches, and even some
recent NAS-based approaches. Additionally, we obtain a
new SOTA result on Flowers-102 with 99.76\% top-1 accuracy, and improve upon the existing baseline on ImageNet
(82.71\% accuracy with 29\% as many parameters as ViT),
as well as NLP tasks. Our simple and compact design for
transformers makes them more feasible to study for those
with limited computing resources and/or dealing with small
datasets, while extending existing research efforts in data
efficient transformers.
\fi


The rest of this paper proceeds as follows. In Sec.~\ref{sec:litreview}, we briefly summarize the related literature on different flood prediction problems. Sec.~\ref{sec:studyarea} describes the considered coastal area and its hydrodynamic model. In Sec.~\ref{sec:approach}, we formalize the prediction problem under study and highlight the associated challenges. Further in Sec.~\ref{sec:approach}, we lay out the proposed Deep Vision-based surrogate modeling framework and the proposed lightweight CNN model CASPIAN.  Sec.~\ref{sec:experiments} provides a thorough evaluation of the developed DL-based flood prediction models and contrasts their performance against that of common approaches. Lastly, Sec.~\ref{sec:conclusion} closes the paper with concluding remarks on the current limitations of the proposed framework and the corresponding extensions for future work.

\section{Literature Review}
\label{sec:litreview}

Data-driven surrogate models have been explored to predict various categories of floods, including pluvial (rainfall-induced), fluvial (riverine), and coastal (often linked to storm surges). From a design perspective, the existing models can be organized under two broad themes: end-to-end and multi-stage. Another layer of differentiation unfolds in the scope of targeted predictive outcomes: the risk, extent, intensity, or dynamics of flooding. In the paragraphs below, we
provide a concise overview of recent flood prediction studies following the above categorization, while for a more exhaustive survey, the reader is referred to \citep{w10111536,Bentivoglio2022DeepDirections,Jones2023,rra.4144}.

In \citep{Chu2020AnDirections}, an ensemble data-driven method for emulation of fluvial inundations was investigated and applied to a river segment in Queensland, Australia. The area of interest was represented as a collection of $14,227$ discrete locations, and for each point, a separate feed-forward type artificial neural network was developed to regress the future state of water depth based on time-series data from past tidal levels and inflows. For pluvial floods, \citep{Guo2021Data-drivenNetworks} designed a CNN-based surrogate modeling approach that supports end-to-end predictions in large-scale urban areas. To cope with the high spatial resolution, the images rasterized from the terrain properties of the catchment areas were split into fixed-size patches, which then, along with hyetographs (rainfall intensities), were fed to the CNN as inputs. The output patches containing the predicted maximum water depths were assembled to form the final inundation map. Expanding the scope further, the work in \citep{Hofmann2021FloodGAN:Time} provided a pluvial flooding metamodel that accommodates spatially non-uniform precipitation events. The metamodel, termed floodGAN, employs a conditional Generative Adversarial Network (cGAN) to predict 2D inundation depth maps directly from raw images of rainfall distribution data. As remarked by the authors, despite the promising predictive capacity of floodGAN, its development and calibration might prove rather complicated and involved, as the process requires simultaneous training of two competing deep neural networks.

In nearshore domains, flood prediction has often been studied under one specific triggering factor (e.g., wind) or within the context of a short-term extreme event (e.g., a storm). For the latter setting, an extensive comparative analysis was provided in \citep{al2020application}, where the authors explored and summarized the frequently employed surrogate modeling techniques in the literature. Three techniques were covered, namely Artificial Neural Networks, Gaussian Process Regression (a.k.a., Kriging), and Support Vector Regression. Based on each, a metamodel was trained to predict storm surge heights at four selected locations with different coastal characteristics, and their performance was assessed. Considering the wind-induced scenario, Luo et al. ~\citep{luo2023wave} implemented two surrogate models, based on a Multi-layer Perceptron and a Decision Tree, to forecast the temporal dynamics of significant wave heights (SWHs), mean wave period and direction for different lead time periods from $12$ to $72$ hours. Both models were trained on hourly wind data from the ERA5 database and applied to the northern part of the South China Sea.

To support high-resolution predictions over extended coastal/urban domains, prior works have alternatively considered two-stage modeling approaches that involve a dimensionality reduction step. For instance, in \citep{Kyprioti2021StormModels}, the landfall locations of storms along the coasts of New Jersey and New York were first grouped with the K-means algorithm, then the hazard curves produced for the centroids of these clusters were interpolated over the original grid via a Kriging-based surrogate model. In \citep{Jia2016SurrogateStorms, Jia2019InvestigationAnalysis, kyprioti2022integration, rohmer2023improved}, different combination strategies of Kriging metamodeling with Principal Component Analysis (PCA) and clustering were explored. The study in~\citep{ElGarroussi2022TacklingFlooding} proposed a two-step surrogate hydraulic model in which the low-dimensional latent representation of the output variables is inferred through an Autoencoder (AE) as opposed to PCA. The results based on AE reduction were found to yield more accurate predictions. Unlike their end-to-end counterparts, however, two-step approaches tend to propagate errors due to their sequential structure, thereby potentially limiting the predictive power of the resulting surrogate model.

With the exception of the work in \cite{Jia2019InvestigationAnalysis}, to the best of our knowledge, flood prediction has not been studied under the joint consideration of future climate change impacts and shoreline armoring scenarios. In~\cite{Jia2019InvestigationAnalysis}, the authors developed a two-stage surrogate hydrodynamic model, based on Kriging and PCA, to investigate the sensitivity of coastal hydrodynamics to shoreline alterations caused by seawall installations under a projected SLR of $1.5$ meters. The model was designed for the county-level protection of the San Francisco Bay area and evaluated (under leave-one-out cross-validation) on a dataset of $40$ scenarios simulated with a high-fidelity hydrodynamic model Delft3D. Here, we approach this coastal flood prediction problem in a different fashion by first reformulating the problem as a computer vision task, then training end-to-end surrogate models that directly support high-resolution predictions. Compared to the aforementioned common two-stage approach of Kriging with PCA, the DL-based surrogate models developed with the proposed strategy confer measurable gains in terms of predictive performance and generalizability, as verified by extensive evaluations reported in Sec.~\ref{sec:experiments}.

\section{Coastal Area and its Hydrodynamic Model}
\label{sec:studyarea}

The geographical area selected for the application of the proposed surrogate modeling framework stretches along the coast of Abu Dhabi, which is the capital of the United Arab Emirates (UAE) situated inside the Persian Gulf. UAE's coastline features a low-lying and shallow-sloping (about $35$ cm per km) topography \citep{melville2021roadmap}. Over $85$\% of the population and more than $90$\% of the local infrastructure of the UAE is within a few meters of the present-day sea level \citep{al2017department}. Notably, Abu Dhabi is comprised of a system of coastal mangrove islands, coral reefs and artificial islands, and $50$\% of its area lies only within $1$ m above sea level \citep{hereher2020assessment,subraelu2022global}. Considering that possible SLR estimates are on the order of $1.0$ to $1.5$ m by the end of $2100$, most of the built and natural mangrove ecosystem of Abu Dhabi, along with its coastal communities, will potentially be exposed to flood hazards and subsequent damages. 

Given the complex structure of Abu Dhabi's coastline, it is necessary to consider the protection of different sections. The partitioning scheme chosen for the purposes of this study was informed by the precincts defined in the 2030 Urban Structure Framework Plan of Abu Dhabi \citep{abu2007plan}. For further refinement of partitions, we divided the main island of Abu Dhabi, grouped other islands, and delineated the boundaries between some precincts, which yielded $17$ individual coastal segments that constitute the candidate sites for installation of engineered fortifications, as depicted in Fig.~\ref{fig:fig1}.  

To allow for detailed and accurate modeling of coastal hydrodynamics of the selected area under SLR, storm events and shoreline fortifications, we adopt a coupled model that combines a Gulf-wide tidal model, a spectral wave model, and a wave run-up model. Fig.~\ref{fig:fig2} demonstrates a typical fit between water level values (relative to the mean sea level) outputted by the validated tidal model and the tidal gauge data at two representative locations near the UAE shore. For illustrations and further details, we refer the reader to Sec.~\ref{sec:AppA} and \citep{chow2022combining}. With this coupled hydrodynamic model in place, one can run a reference case with no shoreline armoring (except those already existing in Abu Dhabi) to evaluate the maximum extent of flooding due to SLR and storms. The placement of protective structures (e.g., seawalls) at candidate coastal sites can be realized by inserting ``fixed weirs'', which will enact flow barriers at the corresponding locations of the domain. For every such configuration of containments, the raw output of the model will include $3$ months worth of hourly water levels for more than $400,000$ grid point locations throughout the Persian Gulf. To filter the nearshore inland locations of interest, based on which the effectiveness of protection scenarios can be appraised, the following two steps were taken: (i) the points lying outside the urban region of Abu Dhabi were excluded; (ii) the inland cells that never experienced flooding even in the case of no coastal protection (i.e., are not hydraulically connected to the Gulf or bear no correlation with the input) were removed. This resulted in the final set of $12066$ locations along the coastline, which appear in Fig.~\ref{fig:fig1}. For each location, the peak water levels (i.e., the maximum value of water depth over the simulated timeframe of $3$ months) under different protection scenarios were then extracted to construct the dataset for training coastal flood prediction models, as elaborated in the following section.


\section{Deep Vision-based Surrogate Modeling Approach}
\label{sec:approach}

In this section, we first define the studied coastal flood prediction problem formally and discuss the associated challenges, then present the details of the proposed surrogate modeling framework, which is graphically summarized in Fig. \ref{fig:fig3}, and the devised minimalistic CNN model CASPIAN, which is illustrated in Fig.~\ref{fig:fig4}.

\noindent{\textbf{Notational Convention:}} Throughout this paper, unless stated otherwise, constants or variables are denoted in normal font (e.g., $H$, $n$), vectors and matrices are distinguished by boldface lowercase and uppercase letters, respectively (e.g., $\bm{x}$, $\bm{X}$), and sets are written in calligraphic or blackboard fonts (e.g., $\mathcal{X}$, $\mathbb{R}$). We let $\bm{0}$ and $\bm{1}$ symbolize the vectors of all zeros and ones, respectively. Lastly, for a given positive integer $n$, the notation $[n]$ shall serve as a shorthand for $\{1, 2, \hdots, n\}$.

\subsection{Problem Statement}
As mentioned in Sec.~\ref{sec:intro}, shoreline adaptations caused by the installation of protective engineering structures (e.g., seawalls) can alter coastal water levels and flood patterns. Specifically, depending on which segments of the coastline these seawalls are raised (i.e., \textit{protection scenario}), the ensuing hydrodynamic interactions and feedbacks can elevate or decrease water levels along other (unprotected) parts of the coast. Accordingly, we focus on the following problem: given an input protection scenario, predict the maximum floodwater levels along the coast. To formalize, denote by $d_{\bm{x}}$ the number of candidate shoreline segments considered for fortification and let $x_i \in \{0,1\}$ be the corresponding decision made for the segment $i \in [d_{\bm{x}}]$ with $1$ indicating the placement of containments and $0$ otherwise. Then, a protection scenario would be represented by a $d_{\bm{x}}$-dimensional binary vector $\bm{x}$ and the set of all possible protection scenarios ($2^{d_{\bm{x}}}$ in total) can be defined as $\mathcal{X} \triangleq \{\bm{x} \mid \bm{x} \in \{0,1\}^{d_{\bm{x}}} \}$. Let $\bm{y}$ be a (non-negative) real-valued vector quantifying the peak water levels at $d_{\bm{y}}$ nearshore locations of interest. With this notation, the prediction problem at hand can be formulated as a regression task of learning a mapping function $f : \bm{x} \in \mathcal{X} \rightarrow \bm{y} \in \mathbb{R}^{d_{\bm{y}}}$ provided with a set $\{(\bm{x}^k, \bm{y}^k) \mid k \in [n], \bm{x}^k \in \mathcal{X}, \bm{y}^k \in \mathbb{R}^{d_{\bm{y}}}\}$ of $n$ available training examples. Since the generation of these input-output pairs involves running high-fidelity hydrodynamic simulations, extensive data collection can prove prohibitively expensive in terms of both time and resources. Consequently, for double-digit values of $d_{\bm{x}}$, the cardinality of the training set can turn disproportionately small compared to that of the input space (i.e., $n << 2^{d_{\bm{x}}}$), enforcing an \textit{extremely low-resource learning setting}. 


The inference of $f$ is further complicated by its output size $d_{\bm{y}}$, which is typically in the order of tens of thousands (here, $d_{\bm{y}} = 12066$). To circumvent this issue, most prior works resort to dimensionality reduction techniques, breaking down the problem into two consecutive steps. First, the original high-dimensional output is transformed into a low-dimensional latent representation, for example through PCA~\citep{Jia2019InvestigationAnalysis,kyprioti2022integration,rohmer2023improved} or an Autoeconder~\citep{ElGarroussi2022TacklingFlooding}, then the surrogate model is trained to map the initial input space to the reduced latent space. Departing from this methodology, we approach the problem through the lens of computer vision and present a strategy for training Deep Vision-based surrogate models capable of producing high-fidelity and high-resolution inundation maps in an end-to-end manner.




\subsection{Proposed Visual Deep Learning Framework}\label{sec:approachvisual}

\begin{figure}[!t]
    \centering
    \includegraphics[width=1\linewidth]{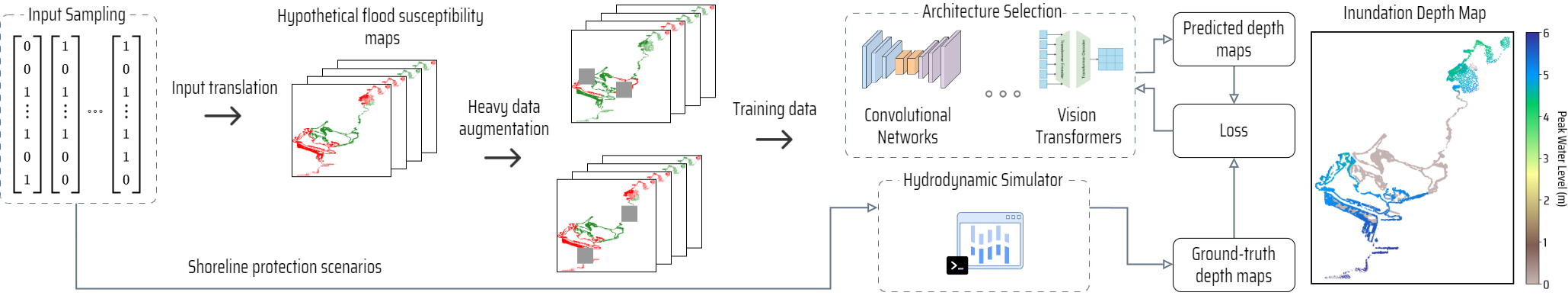}
    \caption{Schematic diagram of the proposed data-driven framework for training performant Deep Vision-based coastal flooding metamodels in low-data settings.} 
    \label{fig:fig3}
\end{figure}

The workflow of the proposed vision-based surrogate modeling framework, displayed in Fig.~\ref{fig:fig3}, can be dissected into four parts, of which first is the generation of training tuples $(\bm{x}^k, \bm{y}^k)$. As the performance of our supervised learning approach hinges on this training data, it is crucial to ensure a sufficiently representative selection of points $(\bm{x}^k)_{k \in [n]}$ for which $f$ will be evaluated, especially under the imposed low-data regime. The scheme adopted herein relies on a combination of judicious manual selection and random sampling. In the former category, the following base scenarios were included: full protection (i.e., $\bm{x} = \bm{1}$), protection of the first and second halves, no protection (i.e., $\bm{x} = \bm{0}$), protection of single precincts (i.e., all binary unit vectors in $\mathcal{X}$) and the inverses thereof, resulting in a total of $4 + 2d_{\bm{x}}$ input instances. For each selected input $\bm{x}^k$, the respective output $\bm{y}^k$ was computed by carrying out a numerical simulation with the coupled hydrodynamic model described in Sec.~\ref{sec:studyarea}.

Recall that every element of $\bm{y}$ corresponds to a specific geographical location parameterized by its latitude and longitude. In vectorial representation, however, this information is abstracted away, leaving the potential of exploiting the spatial correlations and interdependencies between these locations untapped. To enrich the data representation, the proposed pipeline remodels the input and output vectors into matrices as follows. From each $\bm{y}^k, k \in [n]$, we construct a corresponding \textit{flood inundation map} $\bm{Y}^k \in \mathbb{R}^{H \times W}$ through a mapping $\Phi : \mathbb{R}^2 \rightarrow (i,j), i \in [H], j \in [W]$ that converts the geographic coordinates associated with the components of $\bm{y}$ into grid indices $(i,j)$. This transformation $\Phi$ and the grid size $H \times W$ should be selected such that the existing spatial relationships among the output locations are minimally distorted. For the current application site, the coordinate conversion was performed by discretizing the axes of the geographical domain. The dimensions of the formed regular mesh grid, which underlies $\bm{Y}^k$-s, were equated for ease of processing, and the grid size was set as $H \times W = 1024 \times 1024$ to sustain the desired fine geographic granularity of predictions at a reasonable computational cost while maintaining the overall spatial structure of output locations. The mapping conflicts due to discretization were resolved according to the nearest neighbor principle. Subsequently, the established indexing is leveraged to translate the binary protection scenarios $(\bm{x}^k)_{k \in [n]}$ into hypothetical \textit{flood susceptibility maps} $(\bm{X}^k)_{k \in [n]}$, where each $\bm{X}^k \in \mathcal{C}^{H \times W}$ and $\mathcal{C}$ stands for some discrete set of three predefined values that represent categories. Here, the latter was defined\footnote{The choice of values in $\mathcal{C}$ is intended for centering the input data around $0$. Furthermore, for unprotected points one can assign a value proportional to their distance to shoreline rather than a fixed number such as -1, which we leave for future investigations.} as $\mathcal{C} \triangleq \{-1,0,1\}$ and for $\forall~ k \in [n]$, $X_{i,j}^k$ was assigned $-1$ if the shoreline segment in $\bm{x}^k$ closest to the location tied to the $(i,j)$-th index was marked as unprotected, $1$ if protected and the rest of the cells were filled with zeros. In a sense, $\bm{X}^k$-s are segmented matrices in which the $d_{\bm{y}}$ nearshore locations are classified by their distance to unprotected parts of the coast, and the proximity is perceived as a proxy indication of flood risk. It should, nevertheless, be noted that these matrices do not necessarily reflect the actual risk/susceptibility of flooding but are crudely deduced constructs, hence the terming ``hypothetical''. 




Observe that with the remodeled input-output format, the initial regression model is effectively transformed into a problem of learning a mapping of the form $\bm{X} \in \mathcal{C}^{H \times W} \rightarrow \bm{Y} \in \mathbb{R}^{H \times W}$, where $\bm{X}$ and $\bm{Y}$ can be visualized graphically as grayscale (i.e., single channel) images. From a computer vision viewpoint, this problem generally falls under the umbrella of image-to-image translation tasks~\citep{Isola_2017_CVPR}, however, it can also be deemed as a variant of monocular depth estimation from a single image~\citep{yang2024depth, fu2024geowizard} since the predicted output is a depth map (of floodwaters). While both of these directions have been extensively researched under diverse settings, objectives, and applications, to the best of our knowledge, the present problem of inferring depth information from a grayscale, segmented image has not been explored.

Capitalizing on the new image-like representation of inputs and outputs, as a third step of the proposed framework, we artificially increase the volume of training data through image augmentation. Let $\mathcal{D} \triangleq \{(\bm{X}^k, \bm{Y}^k) \mid k \in [n]\}$ be the dataset constructed as prescribed above. From each existing pair $(\bm{X}^k, \bm{Y}^k)$ in $\mathcal{D}$, $m$ new training examples $(\bm{X}^{k(1)}, \bm{Y}^k), \hdots, (\bm{X}^{k(m)}, \bm{Y}^k)$ are generated via the Cutout technique of \citep{devries2017improved}, which applies a fixed-size zero-mask to a random location(s) within the input. By masking out contiguous sections of input images, we essentially erase some information, introducing samples that are partially occluded, noisy copies of original data yet will appear novel to the DL model. Apart from enlarging the size of the training set, this technique, as discussed in \citep{devries2017improved}, exerts a regularization effect, combating the potential for overfitting, and encourages DL models to exploit the full context of the image rather than focus on a few key visual features, which may not always be present. In general, training neural networks on a combination of clean and noisy data, where the noise is added to the inputs, outputs, or gradients, has often been proven instrumental for boosting not only the generalizability but also the predictive capacity of the network, especially in situations when only few training samples are available~\citep{hedderich-klakow-2018-training, goodfellow2016deep}, as is the case here. To further increase the variety of training examples, the Cutout method can be applied in conjunction with other image augmentation techniques, such as rotation, flipping, or shifting. For the sake of the current case study, however, the application of the former method alone (yet in an excessive manner) was found to be sufficient. The size and number of cutout patches, which control the amount of added noise, were determined based on experimentation and are specified in Sec.~\ref{sec:expsetup}.




As the final part of the proposed pipeline, it remains to select the type of neural network that will power the surrogate model and the loss function it will learn to minimize. Now that the problem has been transformed into an image processing task, one has a powerful arsenal of DL techniques at disposal, including both generative models, such as GANs (e.g., pix2pix~\citep{Isola_2017_CVPR}) and Diffusion models (e.g., GeoWizard~\citep{fu2024geowizard}), as well as discriminative models, such as Vision Transformers (e.g., SWIN Transformer ~\citep{liu2021swin}) and CNNs. One salient CNN architecture that has arguably passed the test of time is the U-shaped network, known as U-Net~\citep{unet}, which was originally designed for biomedical imaging tasks, where the available training data is usually scant (as also in the current setting). Since its inception, U-Net has been widely adopted in the biomedical community and beyond, inspiring various new variants and vision models~\citep{azad2022medical}. Drawing on the success of this architecture, we design a minimalistic U-Net-like CNN model aligned to the priorities set forth in this work and the characteristics of the high-resolution flood prediction problem at hand. In particular, we adapt the original U-Net model to (i) further enhance the predictive performance of the network, and (ii) reduce the number of parameters, and hence the \textit{memory and computational resources} required for training, so as to facilitate the \textit{reproducibility} and \textit{accessibility} of the developed coastal flooding metamodel. For exposition clarity, the presentation of the proposed CNN is deferred to the next section. Additionally, we demonstrate the proposed surrogate modeling strategy on two different vision models: a purely transformer-based architecture and a CNN with additive attention gates. The comparison results are reported in Sec.~\ref{sec:experiments}.


Turning to the selection of the loss function, a number of alternatives can be considered, including mean squared Error (\textsc{mse}), mean absolute error (\textsc{mae}), Huber loss~\citep{huber1992robust}, and its reversed variant Berhu~\citep{owen2007robust}. The choice can be informed by analyzing the distribution of water depth values in the dataset and through experimentation. For the current data, the best results were attained with the Huber loss function $L_{\text{Huber}}$, which sets the loss for each point in the output to
\begin{equation}
L_{\text{Huber}} (\delta) = \begin{cases} 
	\frac{1}{2} \delta^2, & \text{If }  |\delta| \leq \theta\\ 
	\theta |\delta| - \frac{1}{2} \theta^2 & \text{otherwise}
\end{cases}\,,
\label{eq:eq1}
\end{equation}
where $\delta$ denotes the error between the predicted and ground truth water depth values and $\theta \geq 0$ is a parameter. When the error doesn't exceed the threshold $\theta$, which was set to $0.5$ in this case study, Huber loss behaves like \textsc{mse}, whereas for larger errors, it mimics \textsc{mae}, thereby enjoying the benefits of both. Recall that by construction, the predicted inundation maps will contain artificially added (background) points for which depth estimation is irrelevant. Therefore, the latter were masked out, and the loss was evaluated only on the valid points that correspond to the $d_{\bm{y}}$ locations of interest.

\subsection{CASPIAN}
\label{sec:CASPIAN}
\begin{figure*}[!t]
     \centering
     \includegraphics[width=\textwidth]{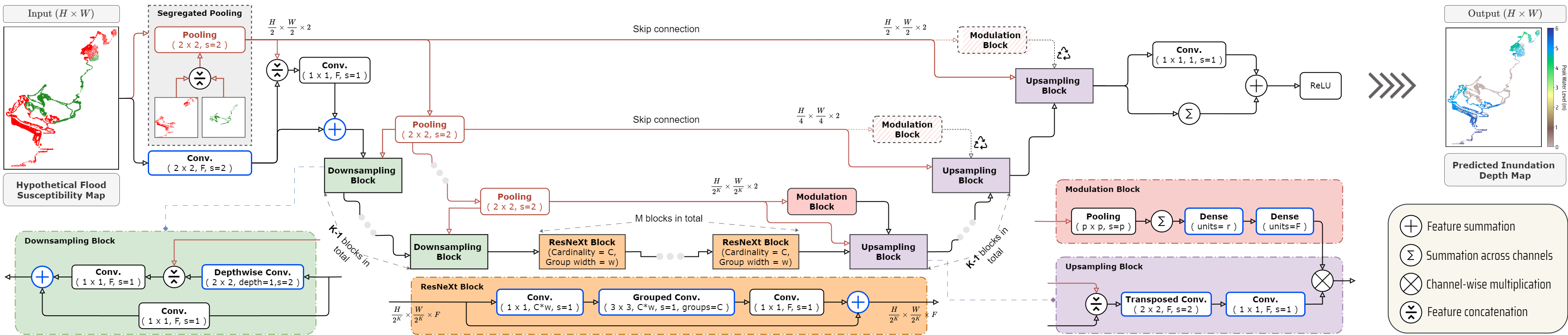}
     \caption{Detailed architecture of the proposed minimalistic CNN model, CASPIAN, for high-resolution coastal flood prediction under SLR and shoreline fortifications. The input image passes through two concurrent paths: a pooling path (colored in red) and a fully convolutional path. The modulation blocks drawn as sketches in dotted outlines are optional and could be substituted by the output from the initial block. The operations followed by non-linear activation functions are marked with a blue border.} 
     \label{fig:fig4}
\end{figure*}

The architecture of the proposed lightweight CNN model CASPIAN, a detailed breakdown of which is presented in Fig.~\ref{fig:fig4}, can be interpreted as a two-layered structure consisting of (i) a fully convolutional encoder-decoder network with a central bottleneck comprised of a series of ResNeXt~\citep{8100117} blocks, and (ii) a cascade of consecutive pooling operations and corresponding supervision blocks linked by skip connections and stacked on top of the encoder and decoder, respectively. The input flood susceptibility maps are simultaneously fed into both of these pathways. While running parallel to one another, these two paths operate in tandem: at every downsampling (upsampling) stage within the network, the outputs from the top pooling path are merged into (multiplied with) the feature maps produced by the bottom convolutional path. The naming of CASPIAN stems from its two distinctive features, namely the cascaded pooling operations and the deep central bottleneck with aggregated residual transformations. The idea behind this dual-path architecture rests on the observation that, under the proposed input representation, the pooling layers, which are traditionally applied after convolutions to compress the extracted feature maps, can instead be employed for capturing the \textit{global context} of the input image, which in our case amounts to the detection of protected and unprotected precincts. In what follows, we discuss the constituents of the proposed model separately, elaborating on their structure, role, and key parameters.

The encoder part of CASPIAN consists of $K$ successive downsampling blocks, which progressively filter and downscale (by a factor of $2$ each) the input image (of size $H \times W$) to generate low-resolution hierarchical feature representations. To allow for efficient utilization of model parameters, we construct these blocks in a style similar to Xception~\citep{8099678}. Specifically, each block, except the first, is built from depthwise convolutions (with stride 2) followed by concatenation (with the feature maps from the pooling path), then pointwise (i.e., $1 \times 1$) convolutions and a residual connection around them. The initial downsampling block, which for clarity is illustrated in a disassembled form in the topmost left corner of Fig.~\ref{fig:fig4}, instead employs a regular convolution with $F$ filters. Typically, in encoder-decoder architectures, the number of feature channels $F$ is doubled at each downsampling step to compensate for the loss
of spatial information caused by compression. Due to their inherently sparse representation, the input flood susceptibility grids lack richness in terms of low-level features (e.g., edges or contours). Therefore, to save the number of parameters at higher resolutions, we keep the number of filters constant across all downsampling blocks. The first block is additionally supplied with the output of a stack of operations from the pooling path, which collectively we refer to as \textit{segregated pooling}. This unit filters the non-background points in the segmented input maps based on their class values into separate channels, which are then concatenated and fed into a pooling layer.

The central segment of CAPSIAN, which serves as a bridge between the encoder and decoder, is formed by $M$ repeated ResNeXt blocks with identical configurations and fixed output size of $\frac{H}{2^K} \times \frac{W}{2^K} \times F$. Each block aggregates identity mapping with a set of transformations realized through grouped and pointwise convolutions, as illustrated in Fig.~\ref{fig:fig4}. As the low-resolution feature maps produced by the encoder undergo these transformations, the proposed network learns more complex and increasingly global (due to enlarging receptive field) feature representations. In addition to the depth $M$, this bottleneck path is parameterized by cardinality $C$ and group width $w$, which control the size and extent of the transformations.

The decoding module in CASPIAN, structurally mirroring the encoder, is assembled from $K$ blocks, which, relying on transposed convolutions (a.k.a., deconvolutions) and pointwise operations, learn to gradually upsample the feature maps distilled by the bottleneck back to the original input resolution $H \times W$. Similar to SegNet~\cite{badrinarayanan2017segnet}, instead of channeling the entire feature maps from the encoder to the decoder through skip connections as in U-Net, the proposed network transfers only the output of corresponding pooling layers as depicted in Fig.~\ref{fig:fig4}. Additionally, these feature maps are reused for modeling the hydrodynamic interactions among protected and unprotected parts of the coast and guiding the decoding process accordingly. In particular, we complement the first upsampling block with a Modulation block constructed similarly to Squeeze-and-Excitation (SE) unit~\citep{8578843}. This block takes the propagated pooling maps as input and produces a set of $F$ weights, one for each channel in the upsampled feature maps. Scaling the latter with these weights allows the network to recalibrate and rectify the decoding process, selectively emphasizing some channels and suppressing others. As illustrated in Fig.~\ref{fig:fig3}, in subsequent upsampling steps, the corresponding modulation blocks can be substituted by the output of the first block. 

The output from the decoder is fed to a $1 \times 1$ convolution and simultaneously summed over channels. The resulting two $H \times W \times 1$ feature maps are summed, and a ReLU activation is applied to it to produce the predicted inundation depth map. The incorporation of the summation operator (which incurs no additional trainable parameters) serves two purposes: (i) aid in preventing overfitting by keeping the learned kernel weights in the final convolutional layers small, and (ii) improve the training by forcing the network to reuse the decoder feature maps (see Sec.~\ref{sec:exprablation} for supporting ablation experiments).
\section{Flood Prediction under SLR and Shoreline Armoring}\label{sec:experiments}

\subsection{Setup and Settings}\label{sec:expsetup}

\noindent{\textbf{Dataset:}} Following the proposed workflow presented in Sec.~\ref{sec:approachvisual}, a total of $142$ input protection scenarios were generated, and the corresponding inundation maps were produced with the adopted physics-based coupled hydrodynamic model. These were randomly distributed into training, validation and testing datasets according to $112$-$12$-$18$ partitioning with no overlaps among the three sets. To ensure the robustness of the results and the reliability of the evaluation, this random process was repeated three times, resulting in three different training, validation and testing datasets. On the training and validation sets, $19$-fold data augmentation was applied through the Cutout technique with two patches, each of size $60 \times 60$. To test the generalizability of the developed models, we additionally construct a Holdout dataset consisting of $32$ handcrafted protection scenarios (see Sec.~\ref{sec:AppB}).

\noindent{\textbf{Candidate Approaches:}} The pool of coastal flood prediction methods selected for evaluation involves two groups of models. The first set includes two standard regression techniques, namely Linear Regression and Lasso Regression with polynomial features (referred to as Lasso with Poly.), and two commonly employed coastal flooding metamodels (as informed by the literature review in Sec.~\ref{sec:litreview}), namely Kriging with PCA and Support Vector Regression (SVR). The second group is composed of four end-to-end DL-based models developed under the proposed framework. Among these, three are based on two existing networks, Attention U-net~\citep{oktay2022attention} and SWIN-Unet~\citep{cao2022swin}, originally designed for medical image segmentation. To adapt to the present settings, the segmentation heads in both networks were replaced by a $1 \times 1$ convolution with a ReLU activation. Additionally, to experiment with the transfer learning technique, we substitute the encoder stack in Attention U-net with the first $16$ convolutional layers from the VGG19 network~\citep{simonyan2014very} and consider two versions, one with the encoder weights initialized randomly while the other with those pre-trained on the popular ImageNet dataset, which contains more than a million images. The latter model is denoted as Attention U-net$^{\ddagger\ddagger}$ to discern between these two. Accordingly, to conform to the three-channel input format of VGG19, for both models, the depth of input matrices was expanded by replacing the class values with RGB codes, resulting in an input size of $H \times W \times 3$. The final fourth model in this cohort was based on the introduced lightweight CNN network CASPIAN. To allow for an impartial inter-group comparison, predictions produced by the models in the first group were \textit{post-processed} to replace the \textit{negative values} with zeros. As an additional reference, we employ a naive regressor, termed Baseline Predictor, which outputs $0$ if the corresponding coastal location in the input vector was (hypothetically) classified as inundation-safe or otherwise the average peak water level across the entire dataset. 

\noindent{\textbf{Model Configurations and Implementation Details:}} Settings of the classical and generalized regression models comprising the first group were determined under experimentation and are listed in Table~\ref{tab:tab2}. Linear Regression, SVR, Lasso with Poly., and Kriging with PCA were implemented via Scikit-learn~\citep{pedregosa2011scikit} and SMT~\citep{SMT2019} Python packages. The implementations (in Tensorflow Keras v2.1) of Attention U-net and SWIN-Unet were borrowed and adapted from~\citep{keras-unet-collection}. The proposed model CASPIAN was built with Tensorflow Keras v2.1. The hyperparameters of Attention U-net were tuned manually and then transferred identically (except the weight initialization in the encoder) to Attention U-net$^{\ddagger\ddagger}$. For SWIN-Unet and CASPIAN, the selection of hyperparameters was optimized through the Random Search algorithm 
provided as part of the Keras Tuner library~\citep{omalley2019kerastuner}. For brevity, Table~\ref{tab:tab2} reports only the total number of trainable parameters of these models, whereas the values of hyperparameters are relegated to Sec.~\ref{sec:AppC}. 

\setlength\extrarowheight{2pt}
\begin{table*}[!b]
\centering
\caption{Quantitative comparison of the candidate coastal flood prediction methods (best viewed in color). The models developed through the proposed Deep Visual Learning framework are highlighted in green. The results for each metric are reported as the mean and standard deviation across the samples over the three data splits. The top scores are highlighted in blue, and the runner-ups are in orange. The superscript $^{\ast\ast}$ denotes an ensemble of individual models, each trained for one specific coastal location.}
\label{tab:tab1}
\resizebox{\textwidth}{!}{%
\begin{tabular}{lcccccccccccccc}
\toprule
\multicolumn{1}{l}{\multirow{5}{*}{\textbf{Model}}} &
  \multicolumn{7}{c}{\multirow{2}{*}{\textbf{Test Dataset (18 samples)}}} &
  \multicolumn{7}{c}{\multirow{2}{*}{\textbf{Holdout Dataset (32 samples)}}} \\
\multicolumn{1}{c}{} &
  \multicolumn{7}{c}{} &
  \multicolumn{7}{c}{} \\ \cmidrule(lr){2-8} \cmidrule(lr){9-15}
\multicolumn{1}{c}{} &
  \multicolumn{5}{c}{\multirow{1}{*}{\textit{( Lower is better )}}} &
  \multicolumn{2}{c}{\multirow{1}{*}{\textit{( Higher is better )}}} &
  \multicolumn{5}{c}{\multirow{1}{*}{\textit{( Lower is better )}}} &
  \multicolumn{2}{c}{\multirow{1}{*}{\textit{( Higher is better )}}} \\ 
\cmidrule(lr){2-6} \cmidrule(lr){7-8} \cmidrule(lr){9-13} \cmidrule(lr){14-15}
\multicolumn{1}{c}{} &
  \textsc{amae} &
  \textsc{armse} &
  \textsc{artae} (\%) &
  $\delta > 0.5$ (\%) &
  $\delta > 0.1$ (\%) &
  $R^2$&
  \textsc{Acc}[0] (\%) &
  \textsc{amae} &
  \textsc{armse} &
  \textsc{artae} (\%) &
  $\delta > 0.5$ (\%) &
  $\delta > 0.1$ (\%) &
  $R^2$ &
  \textsc{Acc}[0] (\%) \\ \midrule
  \rowcolor{Azure2}
\multicolumn{1}{l}{Baseline Predictor} &
  \begin{tabular}[c]{@{}c@{}}1.14\\[-3pt] $\pm$ 0.46\end{tabular} &
  \begin{tabular}[c]{@{}c@{}}2.19\\[-3pt] $\pm$ 0.63\end{tabular} &
  \begin{tabular}[c]{@{}c@{}}58.5\\[-3pt] $\pm$ 32.3\end{tabular} &
  \begin{tabular}[c]{@{}c@{}}27\\[-3pt] $\pm$ 11.2\end{tabular} & 
  \begin{tabular}[c]{@{}c@{}}55.9\\[-3pt] $\pm$ 25.7\end{tabular}&
  \begin{tabular}[c]{@{}c@{}}-0.11\\[-3pt] $\pm$ 0.25\end{tabular} &
  \multicolumn{1}{c}{\begin{tabular}[c]{@{}c@{}}57.9\\[-3pt] $\pm$ 29.9\end{tabular}} &
  \begin{tabular}[c]{@{}c@{}}1.33\\[-3pt] $\pm$ 0.28\end{tabular} &
  \begin{tabular}[c]{@{}c@{}}2.49\\[-3pt] $\pm$ 0.28\end{tabular} &
  \begin{tabular}[c]{@{}c@{}}62.2\\[-3pt] $\pm$ 19\end{tabular} &
  \begin{tabular}[c]{@{}c@{}}29.9\\[-3pt] $\pm$ 7\end{tabular} &  \begin{tabular}[c]{@{}c@{}}52.6\\[-3pt] $\pm$ 12.3\end{tabular}
   &
  \begin{tabular}[c]{@{}c@{}}-0.05\\[-3pt] $\pm$ 0.23\end{tabular} &
  \begin{tabular}[c]{@{}c@{}}70.9\\[-3pt] $\pm$ 12\end{tabular} \\
  \rowcolor{LightYellow2}
\multicolumn{1}{l}{Linear Regression} &
  \begin{tabular}[c]{@{}c@{}}0.20\\[-3pt] $\pm$ 0.06\end{tabular} &
  \begin{tabular}[c]{@{}c@{}}0.57\\[-3pt] $\pm$ 0.15\end{tabular} &
  \begin{tabular}[c]{@{}c@{}}34\\[-3pt] $\pm$ 88.2\end{tabular} &
  \begin{tabular}[c]{@{}c@{}}12.5\\[-3pt] $\pm$ 2.85\end{tabular} & 
  \begin{tabular}[c]{@{}c@{}}19.3\\[-3pt] $\pm$ 4.01\end{tabular}&
  \begin{tabular}[c]{@{}c@{}}0.88\\[-3pt] $\pm$ 0.17\end{tabular} &
  \begin{tabular}[c]{@{}c@{}}41.4\\[-3pt] $\pm$ 17.1\end{tabular} &
  \begin{tabular}[c]{@{}c@{}}0.2\\[-3pt] $\pm$ 0.09\end{tabular} &
  \begin{tabular}[c]{@{}c@{}}0.58\\[-3pt] $\pm$ 0.21\end{tabular} &
  \begin{tabular}[c]{@{}c@{}}9.18\\[-3pt] $\pm$ 2.69\end{tabular} &
  \begin{tabular}[c]{@{}c@{}}12.3\\[-3pt] $\pm$ 3.61\end{tabular} & 
  \begin{tabular}[c]{@{}c@{}}18.6\\[-3pt] $\pm$ 4.12\end{tabular}&
  \begin{tabular}[c]{@{}c@{}}0.93\\[-3pt] $\pm$ 0.05\end{tabular} &
  \begin{tabular}[c]{@{}c@{}}41.8\\[-3pt] $\pm$ 15.6\end{tabular} \\
  \rowcolor{LightYellow2}
\multicolumn{1}{l}{Kriging with PCA} &
  \begin{tabular}[c]{@{}c@{}}0.19\\[-3pt] $\pm$ 0.07\end{tabular} &
  \begin{tabular}[c]{@{}c@{}}0.55\\[-3pt] $\pm$ 0.16\end{tabular} &
  \begin{tabular}[c]{@{}c@{}}22.1 \\[-3pt] $\pm$ 49.5\end{tabular} &
  \begin{tabular}[c]{@{}c@{}}11.3\\[-3pt] $\pm$ 4.2\end{tabular} &
  \begin{tabular}[c]{@{}c@{}}18.4\\[-3pt] $\pm$ 5.26\end{tabular}&
  \begin{tabular}[c]{@{}c@{}}0.91\\[-3pt] $\pm$ 0.09\end{tabular} &
  \multicolumn{1}{c}{\begin{tabular}[c]{@{}c@{}}42.1\\[-3pt] $\pm$ 16.4\end{tabular}} &
  \begin{tabular}[c]{@{}c@{}}0.21\\[-3pt] $\pm$ 0.09\end{tabular} &
  \begin{tabular}[c]{@{}c@{}}0.59\\[-3pt] $\pm$ 0.22\end{tabular} &
  \begin{tabular}[c]{@{}c@{}}9.31\\[-3pt] $\pm$ 3.01\end{tabular} &
  \begin{tabular}[c]{@{}c@{}}12\\[-3pt] $\pm$ 3.83\end{tabular} &
  \begin{tabular}[c]{@{}c@{}}19.3\\[-3pt] $\pm$ 4.51\end{tabular}&
  \begin{tabular}[c]{@{}c@{}}0.93\\[-3pt] $\pm$ 0.05\end{tabular} &
  \begin{tabular}[c]{@{}c@{}}42.8\\[-3pt] $\pm$ 14.6\end{tabular} \\
  \rowcolor{LightYellow2}
  \multicolumn{1}{l}{SVR$^{\ast\ast}$} &
  \begin{tabular}[c]{@{}c@{}}0.17\\[-3pt] $\pm$ 0.08\end{tabular} &
  \begin{tabular}[c]{@{}c@{}}0.69\\[-3pt] $\pm$ 0.26\end{tabular} &
  \begin{tabular}[c]{@{}c@{}}19.9\\[-3pt] $\pm$ 42.1\end{tabular} &
  \begin{tabular}[c]{@{}c@{}}4.47\\[-3pt] $\pm$ 1.68\end{tabular} & 
  \begin{tabular}[c]{@{}c@{}}6.27\\[-3pt] $\pm$ 2.76\end{tabular}&
  \begin{tabular}[c]{@{}c@{}}0.87\\[-3pt] $\pm$ 0.08\end{tabular} &
  \begin{tabular}[c]{@{}c@{}}21.3\\[-3pt] $\pm$ 8.1\end{tabular} &
  \begin{tabular}[c]{@{}c@{}}0.17\\[-3pt] $\pm$ 0.12\end{tabular} &
  \begin{tabular}[c]{@{}c@{}}0.7\\[-3pt] $\pm$ 0.33\end{tabular} &
  \begin{tabular}[c]{@{}c@{}}7.65\\[-3pt] $\pm$ 3.94\end{tabular} &
  \begin{tabular}[c]{@{}c@{}}4.44\\[-3pt] $\pm$ 2.61\end{tabular} & 
  \begin{tabular}[c]{@{}c@{}}6.13\\[-3pt] $\pm$ 3.01\end{tabular}&
  \begin{tabular}[c]{@{}c@{}}0.9\\[-3pt] $\pm$ 0.1\end{tabular} &
  \begin{tabular}[c]{@{}c@{}}19.2\\[-3pt] $\pm$ 6.84\end{tabular} \\
  \rowcolor{LightYellow2}
    \multicolumn{1}{l}{Lasso with Poly.} &
  \begin{tabular}[c]{@{}c@{}}0.14\\[-3pt] $\pm$ 0.05\end{tabular} &
  \begin{tabular}[c]{@{}c@{}}\textbf{\color{Blue3}0.42}\\[-3pt] $\pm$ \textbf{\color{Blue3}0.15}\end{tabular} &
  \begin{tabular}[c]{@{}c@{}}22.5\\[-3pt] $\pm$ 55.6\end{tabular} &
  \begin{tabular}[c]{@{}c@{}}5.45\\[-3pt] $\pm$ 2.65\end{tabular} & 
  \begin{tabular}[c]{@{}c@{}}16.3\\[-3pt] $\pm$ 5.16\end{tabular}&
  \begin{tabular}[c]{@{}c@{}}\textbf{\color{Blue3}0.95}\\[-3pt] $\pm$ \textbf{\color{Blue3}0.03}\end{tabular} &
  \begin{tabular}[c]{@{}c@{}}10.8\\[-3pt] $\pm$ 3.82\end{tabular} &
  \begin{tabular}[c]{@{}c@{}}0.16\\[-3pt] $\pm$ 0.08\end{tabular} &
  \begin{tabular}[c]{@{}c@{}}0.46\\[-3pt] $\pm$ 0.22\end{tabular} &
  \begin{tabular}[c]{@{}c@{}}7.09\\[-3pt] $\pm$ 2.42\end{tabular} &
  \begin{tabular}[c]{@{}c@{}}6.72\\[-3pt] $\pm$ 3.07\end{tabular} & 
  \begin{tabular}[c]{@{}c@{}}14.3\\[-3pt] $\pm$ 4.58\end{tabular}&
  \begin{tabular}[c]{@{}c@{}}0.95\\[-3pt] $\pm$ 0.05\end{tabular} &
  \begin{tabular}[c]{@{}c@{}}10.8\\[-3pt] $\pm$ 4.6\end{tabular} \\
  \rowcolor{myGreen}
\multicolumn{1}{l}{Attention U-net} &
  \begin{tabular}[c]{@{}c@{}}0.11\\[-3pt] $\pm$ 0.07\end{tabular} &
  \begin{tabular}[c]{@{}c@{}}0.64\\[-3pt] $\pm$ 0.26\end{tabular} &
  \begin{tabular}[c]{@{}c@{}}5.56\\[-3pt] $\pm$ 5.73\end{tabular} &
  \begin{tabular}[c]{@{}c@{}}1.9\\[-3pt] $\pm$ 1.35\end{tabular} &
  \begin{tabular}[c]{@{}c@{}}4.04\\[-3pt] $\pm$ 2.4\end{tabular}&
  \begin{tabular}[c]{@{}c@{}}0.89\\[-3pt] $\pm$ 0.08\end{tabular} &
  \multicolumn{1}{c}{\begin{tabular}[c]{@{}c@{}}97.3\\[-3pt] $\pm$ 3.08\end{tabular}} &
  \begin{tabular}[c]{@{}c@{}}0.13\\[-3pt] $\pm$ 0.1\end{tabular} &
  \begin{tabular}[c]{@{}c@{}}0.69\\[-3pt] $\pm$ 0.32\end{tabular} &
  \begin{tabular}[c]{@{}c@{}}5.38 \\[-3pt] $\pm$ 3.59\end{tabular} &
  \begin{tabular}[c]{@{}c@{}}2.29\\[-3pt] $\pm$ 2.08\end{tabular} &
  \begin{tabular}[c]{@{}c@{}}4.71\\[-3pt] $\pm$ 2.79\end{tabular}&
  \begin{tabular}[c]{@{}c@{}}0.9\\[-3pt] $\pm$ 0.08\end{tabular} &
  \begin{tabular}[c]{@{}c@{}}\textbf{\color{DarkOrange3}98.8}\\[-3pt] $\pm$ \textbf{\color{DarkOrange3}1.49}\end{tabular} \\
  \rowcolor{myGreen}
  \multicolumn{1}{l}{Attention U-net$^{\ddagger\ddagger}$} &
  \begin{tabular}[c]{@{}c@{}}0.10\\[-3pt] $\pm$ 0.07\end{tabular} &
  \begin{tabular}[c]{@{}c@{}}0.64\\[-3pt] $\pm$ 0.26\end{tabular} &
  \begin{tabular}[c]{@{}c@{}}5.41\\[-3pt] $\pm$ 5.73\end{tabular} &
  \begin{tabular}[c]{@{}c@{}}1.92\\[-3pt] $\pm$ 1.37\end{tabular} &
  \begin{tabular}[c]{@{}c@{}}\textbf{\color{DarkOrange3}3.95}\\[-3pt] $\pm$ \textbf{\color{DarkOrange3}2.38}\end{tabular}&
  \begin{tabular}[c]{@{}c@{}}0.89\\[-3pt] $\pm$ 0.08\end{tabular} &
  \multicolumn{1}{c}{\begin{tabular}[c]{@{}c@{}}97.2\\[-3pt] $\pm$ 3.12\end{tabular}} &
  \begin{tabular}[c]{@{}c@{}}0.12\\[-3pt] $\pm$ 0.1\end{tabular} &
  \begin{tabular}[c]{@{}c@{}}0.69\\[-3pt] $\pm$ 0.32\end{tabular} &
  \begin{tabular}[c]{@{}c@{}}5.3 \\[-3pt] $\pm$ 3.57\end{tabular} &
  \begin{tabular}[c]{@{}c@{}}2.28\\[-3pt] $\pm$ 2.08\end{tabular} &
  \begin{tabular}[c]{@{}c@{}}\textbf{\color{DarkOrange3}4.52}\\[-3pt] $\pm$ \textbf{\color{DarkOrange3}2.84}\end{tabular}&
  \begin{tabular}[c]{@{}c@{}}0.9\\[-3pt] $\pm$ 0.08\end{tabular} &
  \begin{tabular}[c]{@{}c@{}}98.8\\[-3pt] $\pm$ 1.51\end{tabular} \\
  \rowcolor{myGreen}
\multicolumn{1}{l}{SWIN-Unet} &
  \begin{tabular}[c]{@{}c@{}}\textbf{\color{DarkOrange3}0.07}\\[-3pt] $\pm$ \textbf{\color{DarkOrange3}0.04}\end{tabular} &
  \begin{tabular}[c]{@{}c@{}}\textbf{\color{DarkOrange3}0.42}\\[-3pt] $\pm$ \textbf{\color{DarkOrange3}0.18}\end{tabular} &
  \begin{tabular}[c]{@{}c@{}}\textbf{\color{DarkOrange3}3.21} \\[-3pt] $\pm$ \textbf{\color{DarkOrange3}2.15}\end{tabular} &
  \begin{tabular}[c]{@{}c@{}}\textbf{\color{DarkOrange3}1.37}\\[-3pt] $\pm$ \textbf{\color{DarkOrange3}0.89}\end{tabular} &
  \begin{tabular}[c]{@{}c@{}}7.19\\[-3pt] $\pm$ 3.69\end{tabular}&
  \begin{tabular}[c]{@{}c@{}}\textbf{\color{Blue3}0.95}\\[-3pt] $\pm$ \textbf{\color{Blue3}0.03}\end{tabular} &
  \multicolumn{1}{c}{\begin{tabular}[c]{@{}c@{}}\textbf{\color{DarkOrange3}97.3}\\[-3pt] $\pm$ \textbf{\color{DarkOrange3}2.44}\end{tabular}} &
  \begin{tabular}[c]{@{}c@{}}\textbf{\color{DarkOrange3}0.08}\\[-3pt] $\pm$ \textbf{\color{DarkOrange3}0.06}\end{tabular} &
  \begin{tabular}[c]{@{}c@{}}\textbf{\color{DarkOrange3}0.45}\\[-3pt] $\pm$ \textbf{\color{DarkOrange3}0.21}\end{tabular} &
  \begin{tabular}[c]{@{}c@{}}\textbf{\color{DarkOrange3}3.32}\\[-3pt]  $\pm$ \textbf{\color{DarkOrange3}2.05}\end{tabular} &
  \begin{tabular}[c]{@{}c@{}}\textbf{\color{DarkOrange3}1.70}\\[-3pt] $\pm$ \textbf{\color{DarkOrange3}1.61}\end{tabular} &
  \begin{tabular}[c]{@{}c@{}}7.58\\[-3pt] $\pm$ 3.84\end{tabular}&
  \begin{tabular}[c]{@{}c@{}}\textbf{\color{DarkOrange3}0.95}\\[-3pt] $\pm$ \textbf{\color{DarkOrange3}0.04}\end{tabular} &
  \begin{tabular}[c]{@{}c@{}}98.1\\[-3pt] $\pm$ 3.57\end{tabular} \\
  \rowcolor{myGreen}
\multicolumn{1}{l}{CASPIAN} &
  \begin{tabular}[c]{@{}c@{}}\textbf{\color{Blue3}0.06}\\[-3pt] $\pm$ \textbf{\color{Blue3}0.03}\end{tabular} &
  \begin{tabular}[c]{@{}c@{}}0.46\\[-3pt] $\pm$ 0.18\end{tabular} &
  \begin{tabular}[c]{@{}c@{}}\textbf{\color{Blue3}3.12} \\[-3pt] $\pm$ \textbf{\color{Blue3}1.89}\end{tabular} &
  \begin{tabular}[c]{@{}c@{}}\textbf{\color{Blue3}1.06}\\[-3pt] $\pm$ \textbf{\color{Blue3}0.67}\end{tabular} &
  \begin{tabular}[c]{@{}c@{}}\textbf{\color{Blue3}3.14}\\[-3pt] $\pm$ \textbf{\color{Blue3}1.65}\end{tabular}&
  \begin{tabular}[c]{@{}c@{}}\textbf{\color{DarkOrange3}0.94}\\[-3pt] $\pm$ \textbf{\color{DarkOrange3}0.03}\end{tabular} &
  \multicolumn{1}{c}{\begin{tabular}[c]{@{}c@{}}\textbf{\color{Blue3}98.5}\\[-3pt] $\pm$ \textbf{\color{Blue3}1.84}\end{tabular}} &
  \begin{tabular}[c]{@{}c@{}}\textbf{\color{Blue3}0.06}\\[-3pt] $\pm$ \textbf{\color{Blue3}0.04}\end{tabular} &
  \begin{tabular}[c]{@{}c@{}}\textbf{\color{Blue3}0.45}\\[-3pt] $\pm$ \textbf{\color{Blue3}0.19}\end{tabular} &
  \begin{tabular}[c]{@{}c@{}}\textbf{\color{Blue3}2.80}\\[-3pt] $\pm$ \textbf{\color{Blue3}1.46}\end{tabular} &
  \begin{tabular}[c]{@{}c@{}}\textbf{\color{Blue3}1.01}\\[-3pt] $\pm$ \textbf{\color{Blue3}0.87}\end{tabular} &
  \begin{tabular}[c]{@{}c@{}}\textbf{\color{Blue3}3.99}\\[-3pt] $\pm$ \textbf{\color{Blue3}2.67}\end{tabular}&
  \begin{tabular}[c]{@{}c@{}}\textbf{\color{Blue3}0.96}\\[-3pt] $\pm$ \textbf{\color{Blue3}0.03}\end{tabular} &
  \begin{tabular}[c]{@{}c@{}}\textbf{\color{Blue3}99.1}\\[-3pt] $\pm$ \textbf{\color{Blue3}1.10}\end{tabular} \\ \bottomrule
\end{tabular}%
}
\end{table*}
\setlength\extrarowheight{0pt}

\begin{table}[!t]
  \centering
  \setlength\extrarowheight{2pt}
  \begin{minipage}[b]{0.378\linewidth}
    \centering
    \caption{Addendum to Table~\ref{tab:tab1}.}
    \label{tab:tab2}
\resizebox{0.99\textwidth}{!}{%
\begin{tabular}{lccc}
\toprule
\multicolumn{1}{l}{\multirow{2}{*}{\textbf{Model}}} &
  \multirow{2}{*}{\textbf{Input Size}} &
  \multirow{2}{*}{\textbf{Output Size}} &
  \multirow{2}{*}{\textbf{\begin{tabular}[c]{@{}c@{}}Parameters\\ (\# or settings)\end{tabular}}} \\
\multicolumn{1}{l}{}     &                    &         &     \\ \midrule
\rowcolor{Azure2}Baseline Predictor& $12066$            & $12066$ & $1$ \\
\rowcolor{LightYellow2}
Linear Regression         & $17$ & $12066$ & $17$ \\
\rowcolor{LightYellow2}
Kriging with PCA & $17$ & $12066$ &
  \begin{tabular}[c]{@{}c@{}}Reg. f.~:~linear\\[-2pt] Cor. f. : sq. exp.\end{tabular} \\
\rowcolor{LightYellow2}
SVR$^{\ast\ast}$ & 17 & $1 \cdot (12066)$ & \begin{tabular}[c]{@{}c@{}}Kernel : linear\\[-2pt] $C = 5, \epsilon = 0.05$\end{tabular} \\
\rowcolor{LightYellow2}
Lasso with Poly. & $154$ & $12066$ & \begin{tabular}[c]{@{}c@{}} Inter.\textunderscore only: True\\[-2pt] Degree : 2\end{tabular} \\
\rowcolor{myGreen}
Attention U-net & $H \times W \times 3$ & $H \times W \times 1$ & $\approx$ 12 Mil.\\
\rowcolor{myGreen}
Attention U-net $^{\ddagger\ddagger}$ & $H \times W \times 3$ & $H \times W \times 1$ & $\approx$ 12 Mil. \\
\rowcolor{myGreen}
SWIN-Unet & $H \times W \times 1$ & $H \times W \times 1$ & $\approx$ 8.3 Mil. \\
\rowcolor{myGreen}
CASPIAN & $H \times W \times 1$ & $H \times W \times 1$ & $\approx$ 0.36 Mil. \\ \bottomrule
\end{tabular}%
}
  \end{minipage}
  \hfill
  \setlength\extrarowheight{4.92pt}
  \begin{minipage}[b]{0.61\linewidth}
    \centering
\caption{Results of the ablation studies.}
    \label{tab:tab3}
        \resizebox{0.99\textwidth}{!}{%
        \begin{tabular}{lccccccc}
\toprule
\multicolumn{1}{l}{\multirow{2}{*}{\textbf{Model}}} &
  \multicolumn{3}{c}{\textbf{Test Dataset (18 samples)}} &
  \multicolumn{3}{c}{\textbf{Holdout Dataset (32 samples)}} &
  \multirow{2}{*}{\makecell{\textbf{\# of} \\ \textbf{parameters}}}\\ 
\cmidrule(lr){2-4} \cmidrule(lr){5-7}
  \multicolumn{1}{c}{} &
  \textsc{amae} &
  \textsc{armse}&
  $\delta > 0.1$ (\%) &
  \textsc{amae} &
  \textsc{armse}&
  $\delta > 0.1$ (\%) &
  \\ \midrule
CASPIAN$_{\text{B}}$ &
  \begin{tabular}[c]{@{}c@{}}0.06\\[-3pt] $\pm$ 0.03\end{tabular} &
  \begin{tabular}[c]{@{}c@{}}0.44\\[-3pt] $\pm$ 0.19\end{tabular} & 
  \begin{tabular}[c]{@{}c@{}}3.61\\[-3pt] $\pm$ 2.12\end{tabular} &
  \begin{tabular}[c]{@{}c@{}}0.07\\[-3pt] $\pm$ 0.06\end{tabular} &
  \begin{tabular}[c]{@{}c@{}}0.45\\[-3pt] $\pm$ 0.20\end{tabular} &
  \begin{tabular}[c]{@{}c@{}}6.27\\[-3pt] $\pm$ 10.5\end{tabular} &
  = CASPIAN\\
    CASPIAN$_\Gamma$ &
  \begin{tabular}[c]{@{}c@{}}0.06\\[-3pt] $\pm$ 0.04\end{tabular} &
  \begin{tabular}[c]{@{}c@{}}0.45\\[-3pt] $\pm$ 0.19\end{tabular} & 
  \begin{tabular}[c]{@{}c@{}}5.62\\[-3pt] $\pm$ 4.91\end{tabular} &
  \begin{tabular}[c]{@{}c@{}}0.07\\[-3pt] $\pm$ 0.05\end{tabular} &
  \begin{tabular}[c]{@{}c@{}}0.47\\[-3pt] $\pm$ 0.20\end{tabular} &
  \begin{tabular}[c]{@{}c@{}}5.59\\[-3pt] $\pm$ 4.41\end{tabular} &
  $\approx$ 0.215 Mil.\\
  CASPIAN$_{\text{Z}}$ &
  \begin{tabular}[c]{@{}c@{}}0.10\\[-3pt] $\pm$ 0.07\end{tabular} &
  \begin{tabular}[c]{@{}c@{}}0.58\\[-3pt] $\pm$ 0.30\end{tabular} & 
  \begin{tabular}[c]{@{}c@{}}7.04\\[-3pt] $\pm$ 3.40\end{tabular} &
  \begin{tabular}[c]{@{}c@{}}0.11\\[-3pt] $\pm$ 0.09\end{tabular} &
  \begin{tabular}[c]{@{}c@{}}0.62\\[-3pt] $\pm$ 0.29\end{tabular} &
  \begin{tabular}[c]{@{}c@{}}7.46\\[-3pt] $\pm$ 2.81\end{tabular} &
  $\approx$ 0.344 Mil.\\
  CASPIAN$_{\Omega}$ &
  \begin{tabular}[c]{@{}c@{}}0.10\\[-3pt] $\pm$ 0.07\end{tabular} &
  \begin{tabular}[c]{@{}c@{}}0.55\\[-3pt] $\pm$ 0.31\end{tabular} & 
  \begin{tabular}[c]{@{}c@{}}6.21\\[-3pt] $\pm$ 2.97\end{tabular} &
  \begin{tabular}[c]{@{}c@{}}0.11\\[-3pt] $\pm$ 0.09\end{tabular} &
  \begin{tabular}[c]{@{}c@{}}0.62\\[-3pt] $\pm$ 0.29\end{tabular} &
  \begin{tabular}[c]{@{}c@{}}6.93\\[-3pt] $\pm$ 2.81\end{tabular} &
  $\approx$ 0.341 Mil.\\
\bottomrule
\end{tabular}%
}
  \end{minipage}
\end{table}
\setlength\extrarowheight{0pt}

\noindent{\textbf{Training Specifications:}} All the four DL models were trained with Adam optimizer under the Huber Loss (defined in Eq.~\ref{eq:eq1}) function with $\theta = 0.5$ and batch size of  $2$. The adopted learning schedule, determined through trials with several alternatives, was set to start with a gradual warm-up that increases the learning rate from $0$ to $\text{LR}$ linearly for $20$ epochs, followed by $200$ epochs of the main training session wherein the learning rate was reduced ($\times 0.85$) whenever the validation loss plateaued (patience = $10$). During the main training, early stopping was applied if no improvement in the validation loss was recorded for $40$ consecutive epochs. $\text{LR}$ was set to $1.5 \cdot 10^{-4}$ for Attention U-Net and Attention U-net$^{\ddagger\ddagger}$, to $1.8 \cdot 10^{-4}$ for SWIN-Unet, and to $8 \cdot 10^{-4}$ for CASPIAN. All the models were trained and evaluated on a desktop machine with an Intel Core i9 3.00 GHz CPU, 64 GB of RAM and a single NVIDIA RTX 4090 GPU.

\noindent{\textbf{Evaluation Metrics:}}
As emphasized in~\citep{al2020application}, in most prior works studying coastal flood prediction, the performance of developed coastal metamodels was assessed considering only a few basic aggregate metrics, such as \textsc{mse}, \textsc{mae} or $R^2$, which may not adequately reflect the actual quality of predictions. To provide a more comprehensive evaluation, we consider $6$ different metrics, including both error and accuracy measures, which are formalized as follows:
\begin{equation}
\begin{array}{ll}
\textsc{artae} \triangleq \frac{1}{N}\sum_{k=1}^{N} \frac{\| \bm{y}^k - \hat{\bm{y}}^k \|_1}{\| \bm{y}^k \|_1} & R^2 \triangleq \frac{1}{N} \sum_{k=1}^{N} \left( 1 - \frac{\sum_{i=1}^{d_{\bm{y}}} (y_i^k - \hat{y}_i^k)^2}{\sum_{i=1}^{d_{\bm{y}}} (y_i^k - \bar{y}^k)^2} \right)\\
\textsc{armse} \triangleq \frac{1}{N}\sum_{k =1}^{N}\sqrt{\sum_{i=1}^{ d_{\bm{y}}}\frac{(y^k_i - \hat{y}^k_i)^2}{d_{\bm{y}}}} & \delta > \Delta \triangleq \frac{1}{N} \sum_{k=1}^{N} \frac{\big|\{ i ~:~  |y_i^k - \hat{y}_i^k| > \Delta,~ i \in [d_y]\}\big|}{d_{\bm{y}}} \\
\textsc{amae} \triangleq \frac{1}{N}\sum_{k =1}^{N}\sum_{i=1}^{ d_{\bm{y}}}\frac{|y^k_i - \hat{y}^k_i|}{d_{\bm{y}}} & \textsc{Acc}[0] \triangleq \frac{1}{N} \sum_{k=1}^{N} \frac{\big|\{ i ~:~  y_i^k =0,~ i \in [d_y]\}\big|}{d_{\bm{y}}}
\end{array},
\end{equation}
where $N$ is the number of evaluated samples; $\bm{y}$ and $\bm{\hat{y}}$ correspond to the ground truth and predicted peak water levels, respectively, of the $d_y$ locations of interest filtered from the predicted inundation maps; and $\Delta$ is an error threshold (in meters).

\subsection{Results and Comparison}\label{sec:expresults}

Table~\ref{tab:tab1} summarizes the performance of the candidate models on the Test and Holdout datasets averaged over the three data splits. Notable observations from the first group of models are as follows. The two-stage Kriging with PCA approach, commonly employed in prior works, slightly improves upon Linear Regression by achieving 22.1\% \textsc{artae}, but $\delta$ errors are comparable, indicating a similar prediction quality. On the other hand, we observe significant improvement with Lasso with Poly, which reached the highest \textsc{armse} of 0.42\% and $R^2 = 0.95$ across all the models while achieving a $\delta > 0.5$ error of 5-7\%, about half of the error produced by Kriging with PCA. This signifies that incorporating engineered input features which model the interactions between the protected and unprotected precincts can improve the prediction quality. Among the first group of models, SVR achieved the lowest $\delta$ errors, which indicates a higher quality of predictions, yet \textsc{acc}[0] is extremely low at around 20\%. However, it should be noted that the SVR model requires training a separate model for every output coastal location independently, which raises potential scalability issues.



The DL models in the second group, which were trained with the proposed framework, significantly outperformed the first group in terms of \textsc{amae} (by a factor of 2 on average), \textsc{artae} (by a factor of 2-5) and $\delta$ errors (by a factor of 2-5), and especially for \textsc{acc}[0] (more than two-fold). The version of Attention U-net with the pre-trained weights achieved only modest improvements over the version trained from scratch, specifically a marginal improvement of 0.1\% of \textsc{artae} and 0.01\% of $\delta > 0.5$  error. This result could possibly be attributed to the stark difference in image modalities and sizes between the current dataset and ImageNet, which is consistent with the findings and conclusions drawn in~\citep{9134370}. The two best-performing candidate models were SWIN-Unet and CASPIAN, the former a close runner-up to the latter for the majority of the metrics. Notably, CASPIAN attained the highest scores for all metrics in the Holdout dataset while requiring only a fraction of SWIN-Unet's model's size. As reported in Table~\ref{tab:tab2}, CASPIAN achieved $\delta > 0.5$  error of only 1\% on both datasets, and the performance with respect to the other metrics is consistent on both datasets, demonstrating the generalizability of the proposed CNN model.

\subsection{Ablation Experiments}\label{sec:exprablation}

To supplement the evaluation results reported in Sec.~\ref{sec:expresults}, this section ablates the key architectural components introduced in CASPIAN. In particular, we remove/truncate the blocks/modules individually in four steps, resulting in the following versions: (i) CASPIAN without the final channel-wise summation, abbreviated as CASPIAN$_{\text{B}}$, (ii) CASPIAN with the depth of the central bottleneck reduced to 2 (i.e., $M=2$), denoted as CASPIAN$_\Gamma$, (iii) CASPIAN with the modulation block removed, denoted as CASPIAN$_{\text{Z}}$, (iiii) CASPIAN with the pooling path completely eliminated, referred to as CASPIAN$_{\Omega}$. Since the observed performance of CASPIAN was consistent across the data splits, the ablation studies were performed on one split. 

Table~\ref{tab:tab3} reports the changes to several performance metrics resulting from the ablation experiments. CASPIAN$_{\text{B}}$ and CASPIAN$_\Gamma$ achieved similar results on the Test dataset compared to CASPIAN, yet $\delta > 0.1$ error of the produced predictions on the Holdout dataset nearly doubled, indicating poorer generalizability. This observation corroborates the importance of the proposed deep central bottleneck and the final summation operator. In the case of CASPIAN$_{\text{Z}}$ and CASPIAN$_{\Omega}$, a significant drop in performance was observed on both datasets approaching that of Attention U-Net. This outcome can be expected since, after the elimination of the pooling path, the architecture of the latter two more closely resembles that of Attention U-Net.

\section{Concluding Remarks}
\label{sec:conclusion}

In this paper, we presented a data-driven surrogate modeling framework for developing accurate and reliable coastal flooding metamodels powered by vision-based DL techniques. The proposed framework was tested on three different DL architectures, including a lightweight CNN model CASPIAN introduced in this work. The developed models were shown to significantly outperform existing geostatistical methods and standard regression techniques commonly employed in prior studies. The best-performing model, CASPIAN, closely and consistently emulated the results of the high-fidelity hydrodynamic simulator, on average achieving an $\textsc{amae}$ error of $0.06$ and $\delta > 0.5$ error of only around 1\% on both Test and Holdout datasets.  
 
One limitation of the developed coastal metamodels is that they are currently domain-specific since the training was performed considering one specific shoreline, SLR scenario, and fixed set of wind parameters. Without major modifications to the proposed framework, one can extend the proposed framework to account for different coastal areas, for example, by including other geographical data such as local slopes, and hydraulic connectivities in the input maps. Another possible extension would be to expand the predictive scope and, in addition to peak water levels, also estimate the maximum velocities of floodwaters, a key prediction needed for coastal damage assessment.

\bibliographystyle{icml2024}
\bibliography{references, references_misc}

\appendix

\section{Further Details on the Adopted Hydrodynamic Model}
\label{sec:AppA}

\begin{figure}[!b]
    \centering
    \includegraphics[width=\linewidth]{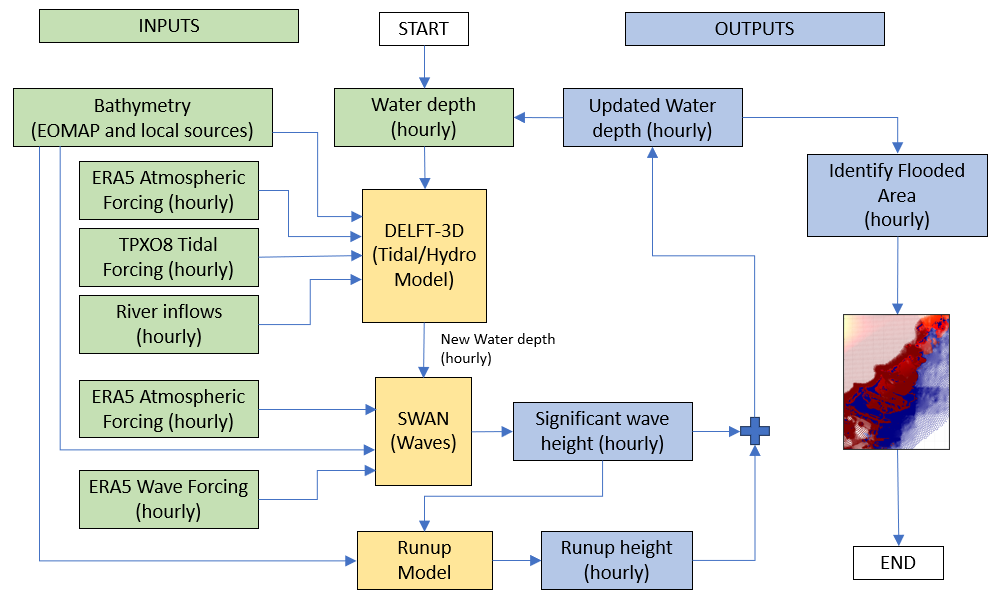}
    \caption{Schematic of the adopted coupled hydrodynamic model. Green elements denote input parameters; Yellow are the constituent sub-models; Blue are model outputs. Running one cycle of the above model will generate an hourly update of the water depth throughout the UAE coastline.}
    \label{fig:Hydro_Schematic}
\end{figure}

As noted in Sec.~\ref{sec:studyarea}, the adopted model combines a Gulf-wide hydrodynamics model (Delft3D\footnote{https://oss.deltares.nl/web/delft3d}), a spectral wave model (SWAN\footnote{https://swanmodel.sourceforge.io}), and a wave run-up model to predict inundated parts of Abu Dhabi's coastline resulting from SLR and storm events. Fig.~\ref{fig:Hydro_Schematic} provides a schematic of the inputs and outputs of the three models. In brief, at each hour of the simulation period, the Delft3D model takes in multiple inputs (such as tidal and atmospheric forcing, plus local bathymetry) and outputs a water level for the SWAN model, which in turn provides a significant wave height for the run-up model. The significant wave heights are then fed back to the Delft3D for the next time step. More details of each model can be found in \cite{chow2022combining}.

Delft3D is a hydrodynamic model that solves the time-dependent Reynolds Averaged Navier Stokes differential equations. That is, Delft3D is a physics-based numerical model that considers the time-varying forces exerted on a water body (such as the entire Persian Gulf) due to hydrostatic pressures (such as SLR), tidal forcing, wind and storm stresses, bottom (seabed) friction, and river inflows over a finite-element computational grid (up to $30$ m in horizontal resolution) spread over variable bathymetry. For any point in this grid, it can provide time series outputs, with $30$-minute intervals, of water levels and local water circulation velocities throughout the specified simulation period. Importantly, Delft3D can handle computational grid cells that alternate between dry and wet states~\citep{barnard2014development}.

The tidal model was validated by running the Delft3D simulator over a 3-month period between 1 January and 31 March 2017 (without wind forcing) and computing the root mean squared error between the model outputs of hourly water levels at 194 locations throughout the gulf and hourly tidal gauge water level data obtained from the TPXO8 Ocean Atlas for the same period \footnote{https://www.tpxo.net/global/tpxo8-atlas. Also see \citep{egbert2002efficient}}. The model was calibrated by adjusting the bottom Manning’s roughness coefficient for the entire gulf domain from $0.015$ to $0.030$. The lowest overall error was attained under the coefficient of $0.02$, which was taken as the calibrated roughness value for the gulf model going forward. Fig.~\ref{fig:fig2} demonstrates a typical fit between water level values (relative to the mean sea level) outputted by the model and the tidal gauge data at two representative locations near the UAE shore.

To account for wind-induced wave activity in the vicinity of Abu Dhabi's coast, the validated Delft3D model was rerun with wind forcing from the ERA5 database, and the results were fed to an additional spectral wave model, SWAN, which allows capturing wind-wave generation, wave diffraction, amplification and refraction of water surface waves as they approach the shoreline. The SWAN model was applied at a scale of about $100$ km along the shoreline to about $50$ km offshore under the same forcing from the ERA5 database. Finally, along the interface of the waves with the coastline, the SWAN-computed significant wave heights and the local shoreline slope were used to compute the run-up elevations along the coastline where the waves hit the shore.

\section{Holdout Dataset Scenarios}
\label{sec:AppB}

\begin{table}[!h]
\centering
\caption{Protection scenarios considered in the Holdout dataset.}
\resizebox{\textwidth}{!}{%
\begin{tabular}{*{5}{c}}
\toprule  
\multicolumn{5}{c}{\textbf{Holdout protection scenarios} (32 in total)} \\
\midrule  
00110011001100110 & 11100000000000111 & 00000111100000111 & 00011000110001100 & 11110000111100001 \\
00000011111100000 & 11110000000001111 & 00000111111100000 & 11111100000111111 & 00001111111110000 \\
11111000001111100 & 00001110000111000 & 10101010101010101 & 11111110000001111 & 00000001111110000 \\
11111000000011111 & 11111110000000111 & 00000111110000011 & 00011100011100011 & 00000001111111000 \\
11000000000000011 & 00111111111111100 & 01010101010101010 & 11111100000011111 & 11111000011111000 \\
00000011111000000 & 11110001111000111 & 11100011100011100 & 00001111000011110 & 11001100110011001 \\
11100111001110011 & 00011111111111000 &                   &                   &                   \\
\bottomrule  
\end{tabular}%
}
\end{table}

\section{Hyperparameters of the Trained DL Models}
\label{sec:AppC}

\paragraph{Attention U-net:} 
\begin{itemize}
\item \verb=input_size = = (1024, 1024, 3)
\item \verb=filter_num = = [32, 64, 128, 256] (number of filters for each down- and up-scaling level) 
\item \verb=stack_num_down= = 2 (number of layers per downsampling level/block)  
\item \verb=stack_num_up= = 2 (number of layers (after concatenation) per upsampling level/block 
\item \verb=activation= = 'ReLU'
\item \verb=atten_activation= ='ReLU'
\item \verb=output_activation= ='ReLU'
\item \verb=batch_norm= = False
\item \verb=backbone= = 'VGG19' (the backbone model name)
\item \verb=encoder_weights= = 'random'
\item \verb=freeze_backbone= = False
\end{itemize}

\paragraph{Attention U-net$^{\ddagger\ddagger}$:} 
\begin{itemize}
\item \verb=input_size= = (1024, 1024, 3)
\item \verb=filter_num= = [32, 64, 128, 256]
\item \verb=stack_num_down= 2 
\item \verb=stack_num_up= 2
\item \verb=activation= = 'ReLU'
\item \verb=atten_activation= = 'ReLU'  
\item \verb=output_activation= = 'ReLU'
\item \verb=batch_norm= = False
\item \verb=backbone= = 'VGG19' 
\item \verb=encoder_weights= = 'imagenet' (pre-training on ImageNet), 
\item \verb=freeze_backbone= = False
\end{itemize}

\paragraph{Swin-Unet:}
\begin{itemize}
\item \verb=filter_num_begin= = 64 (number of channels in the first downsampling block)
\item \verb=depth= = 4 (the depth of Swin U-net, 4 means 3 down/upsampling levels and a bottom level) 
\item \verb=stack_num_down= = 2 (number of Swin Transformers per downsampling level)
\item \verb=stack_num_up= = 2 (number of Swin Transformers per upsampling level) 
\item \verb=patch_size= = 8
\item \verb=att_heads= = 4 (number of attention heads per down/upsampling level)
\item \verb=w_size= = 4 (the size of attention window per down/upsampling level)
\item \verb=mlp_ratio= = 4 (ratio of MLP hidden dimension to embedding dimension)
\end{itemize}

\paragraph{CASPIAN:} 
\begin{itemize}
\item \verb=input_shape= = (1024, 1024, 1) 
\item \verb=F= = 72
\item \verb=K= = 4 
\item \verb=C= = 34 
\item \verb=M= = 8 
\item \verb=modulation_level= = 1
\item \verb=p= = 4
\item \verb=r= = 0.85 $\times$ F
\item \verb=activation= = \verb='tanh'= 
\item \verb=init= = \verb="glorot_normal"=
\end{itemize}

\end{document}